\documentclass{article}
\usepackage{blindtext}
\usepackage{orcidlink,thumbpdf,lmodern}
\usepackage{amsmath}
\usepackage{amsfonts}
\usepackage{moresize}
\usepackage{graphicx}
\usepackage{float}
\usepackage{caption}
\usepackage{lipsum}
\usepackage{bm}
\usepackage{mathrsfs}
\usepackage{tikz}
\usetikzlibrary{positioning}
\usepackage{amsthm}
\usepackage{hyperref}

\DeclareMathOperator*{\argmin}{arg\,min}

\title{Introducing Feature-Based Trajectory Clustering, a clustering algorithm for longitudinal data}
\author{Marie-Pierre Sylvestre and Laurence Boulanger}

\begin{document}
\maketitle
\begin{abstract}
We present a new algorithm for clustering longitudinal data. Data of this type can be conceptualized as consisting of individuals and, for each such individual, observations of a time-dependent variable made at various times. Generically, the specific way in which this variable evolves with time is different from one individual to the next. However, there may also be commonalities; specific characteristic features of the time evolution shared by many individuals. The purpose of the method we put forward is to find clusters of individual whose underlying time-dependent variables share such characteristic features. This is done in two steps. The first step identifies each individual to a point in Euclidean space whose coordinates are determined by specific mathematical formulae meant to capture a variety of characteristic features. The second step finds the clusters by applying the Spectral Clustering algorithm to the resulting point cloud.
\end{abstract}

\section{Introduction}
The present paper is the first in a series of 3 companion papers about Feature-Based Trajectory Clustering (FBTC), a new algorithm for clustering continuous and ordinal longitudinal data, henceforth refered to as trajectories. Underlying every trajectory is a time-dependent variable, a function, which has a well defined value at every moment in time but that is unknown, except at those specific times which make up the trajectory. Let us take a moment to anchor these notions in a concrete example. At every moment in time, an individual has a well defined hemoglobin level in their blood, measured in gram per deciliter. Most of the time, this level is unknown to them, except when it is measured by a nurse, roughly every couple of weeks, when they donate blood. Specifically, let us assume that the individual went to donate blood at times $t=0,14,30,43$ (measured in days) and that they recorded their corresponding hemoglobin levels on a note pad as $y_0=15$, $y_{14}=15.4$, $y_{30}=13.9$, $y_{43} = 14.6$. In this scenario, the underlying function $f(t)$ is the level of iron in the individual's blood, and the trajectory is the 4-tuple $(y_0,y_{14},y_{30},y_{43})$. Assuming no error on the part of the nurse and their instrument, the relation between the trajectory and the underlying function is $y_0 = f(0)$, $y_{14}=f(14)$, $y_{30}=f(30), y_{43} = f(43)$. Nothing else is known about $f(t)$.

Given a set of trajectories for various individuals, we are concerned with the problem of identifying clusters of individuals whose underlying functions share common characteristic features. Since the underlying functions themselves are unknown, we are required to work at the level of the trajectories and to assume that the observation times were sufficiently frequent that all of the essential features of the underlying functions that we care about are reflected to an adequate degree by the trajectories. The basic idea goes back to work by Leffondré et al. \cite{L}. For each trajectory, twenty so-called {\em measures} are computed. These are numerical quantities deriving from mathematical formulae meant to capture various features and characteristic behaviors of a trajectory. In this way, the set of trajectories becomes identified to a cloud of points in twenty dimensional euclidean space and can be clustered using one of the preexisting algorithm for clustering data of this type. As a default, we precognize a version of Spectral Clustering for its ability to detect non convex clusters. We note that this way of clustering trajectories is different in its scope to existing methods, like $K$-means and latent class models, which focus on a forming clusters of trajectories based on proximity alone.

Section \ref{section_measures} presents in detail each of the twenty measures and provides tips for interpreting them. Section \ref{section_clustering} proposes an algorithm for clustering the measures and discussed various aspects and properties of the method as a whole. Section \ref{section_examples} provides examples illustrating how the method performs on various data sets.

%
%

\section{The measures}\label{section_measures}
\begin{figure}[!ht]
	\centering
	\includegraphics[scale=0.4]{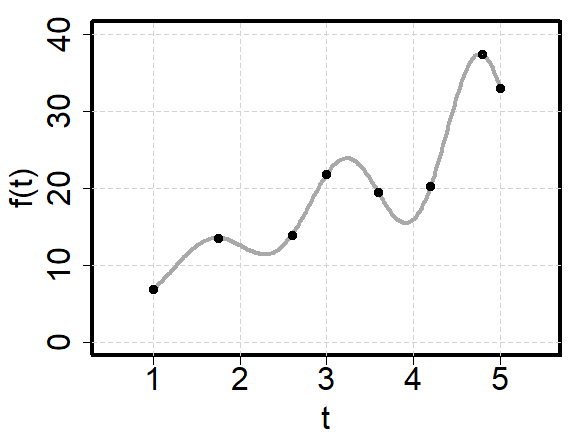}
	\caption{{\footnotesize A trajectory of lenght $N=8$ and its underlying function $f(t)$. The observations times are $t_1=1$, $t_2=1.75$, $t_3=2.6$, $t_4=3$, $t_5=3.6$, $t_6=4.2$, $t_7=4.8$, $t_8=5$.}}
	\label{fig:fct_plus_traj}
\end{figure}

In this section, let $\bm{y}=(y_{1},\ldots,y_{N})$ denote a {\em trajectory} of lenght $N$ with corresponding observation times $\bm{t} = (t_{1},\ldots,t_{N})$, $a\leq t_{1}<\ldots<t_{N}\leq b$ and {\em underlying function} $f\in C^2[a,b]$. This means that $f$ is defined on $[a,b]$, has a continuous second derivative and that the trajectory is the result of evaluating $f$ at the observations times, namely $y_j=f(t_j)$. We do not assume that the times $t_j$ are equidistant from one another. The situation is depicted graphically in Figure \ref{fig:fct_plus_traj}.
Let us stress that, in practical applications, the underlying function $f$ is unknown, but we assume that it exists, just as we assume, unless stated otherwise, that the trajectories are observations of $f$ made without measurement error. We will refer to a function $m[\bm{y},\bm{t}]$ of the trajectory and observation times as a {\em  trajectory measure}. Since our fundamental goal is to cluster trajectories on the basis of their underlying functions, we insist that each trajectory measure should approximate a corresponding {\em functional measure} defined at the level of the underlying function.

\begin{figure}[!ht]
	\centering
	\includegraphics[scale=0.425]{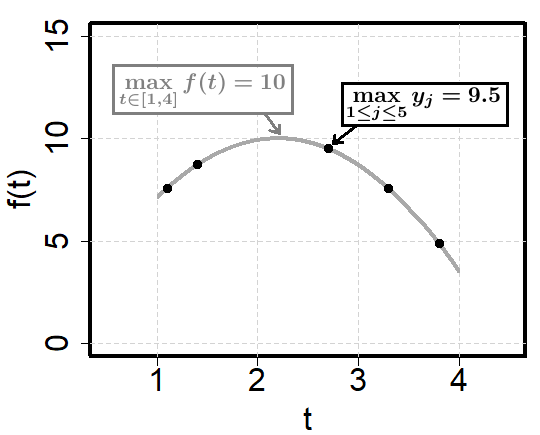}
	\caption{{\footnotesize In this example, the maximum of the the underlying function (in grey) is 10 so the value of the functional measure is 10. The underlying has been observed five times, resulting in a trajectory (black points) of lenght 5. The maximum value of the trajectory is 9.5 so the value of the trajectory meansure is 9.5.}}
	\label{fig:grand_example}
\end{figure}

Our purpose in this section is to present 20 trajectory measures that represent the trajectories and that we will use in section \ref{section_clustering} to form clusters of trajectories. Before going further, let us pause to look at a simple example of a trajectory measure, the maximum. In this case, the functional measure we seek to approximate is $\max_{t\in[a,b]} f(t)$, the maximum value taken by the underlying function on its domain. An appropriate trajectory measure approximating this functional measure is $\max_{1\leq j\leq N} y_j$, the maximum value of the trajectory. See Figure \ref{fig:grand_example} to see how this works on a concrete example.

\begin{figure}[!ht]
	\centering
	\includegraphics[scale=0.4]{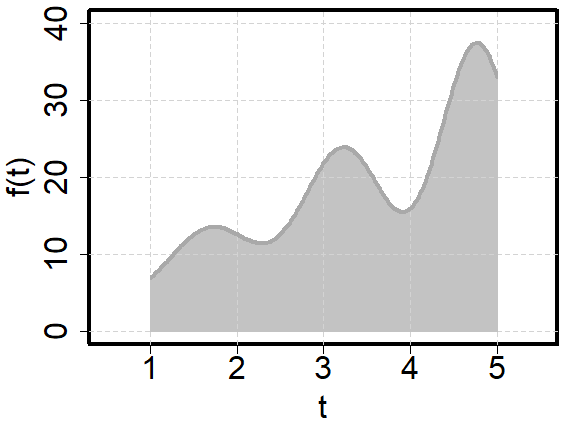}\includegraphics[scale=0.4]{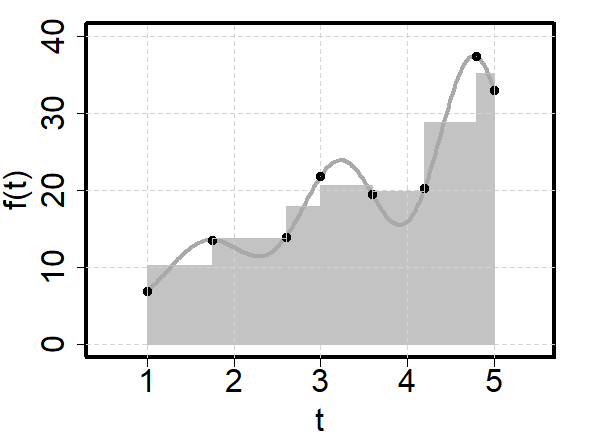}
	\caption{{\footnotesize On the left, the area under the curve (in grey) represents the value of the integral $\int_1^5f(t)\,dt$. On the right, this integral is approximated from a trajectory using the trapezoid rule of numerical integration. For any two consecutive points $(t_j,y_j)$, $(t,_{j+1},y_{j+1})$, a rectangle is constructed with base equal to the distance $t_{j+1}-t_j$ between the observation times and heigh equal to the average of the $y$ coordinates, $(y_j+y_{j+1})/2$. The trapezoidal approximation to $\int_1^5f(t)\,dt$ is the sum of the areas of the 7 rectangles constructed this way. In the present case, the true value of the integral is 75.2 and the trapezoidal approximation is 75.1.}}
	\label{fig:integration}
\end{figure}

Some of the functional measures we define below involve definite integrals of functions. Whenever this is the case, the trajectory measure will be obtained by approximating these integrals using the trapezoidal rule of numerical integration (Figure \ref{fig:integration}).
This means that if $G(t)$ is a function observed at times $t_1<\ldots<t_N$, then
\begin{equation}\label{trapezoidal_rule}
\int_{t_1}^{t_N}G(t)\,dt\approx \sum_{j=1}^{N-1}\frac{G(t_{j}) + G(t_{j+1})}{2}(t_{j+1}-t_j).
\end{equation}
The right hand side of this equation corresponds to taking the mean of the left and right  Riemann sums. Typically, the greater the lenght of the trajectory, the better the approximation. Recall that the integral of a function represents the signed area under its graph, where the word signed means that areas above the $t$ axis are counted positively but areas under the $t$ axis are counted negatively. 

Some of the functional measures we consider involve the first or second derivative of the underlying function. If a functional measure involves the first derivative $f'(t)$ (resp. second derivative $f''(t)$), the corresponding trajectory measure will be obtained by replacing the value of $f'(t)$ (resp. $f''(t)$) at $t=t_j$ with an approximation $D_j$ (resp. $D_j^2$) computed using the trajectory. See appendix A for the details of this approximation.

\begin{figure}[!ht]
	\centering
	\includegraphics[scale=0.4]{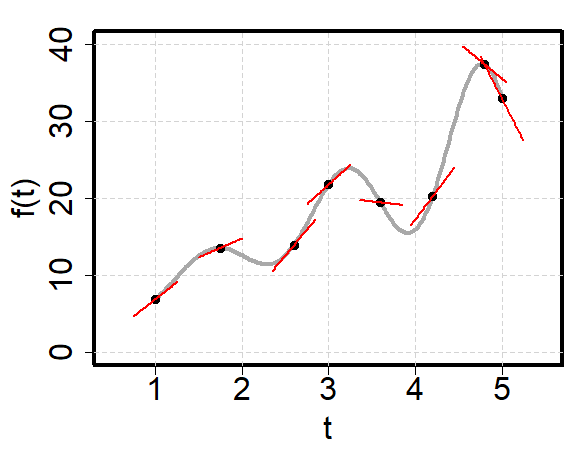}\includegraphics[scale=0.4]{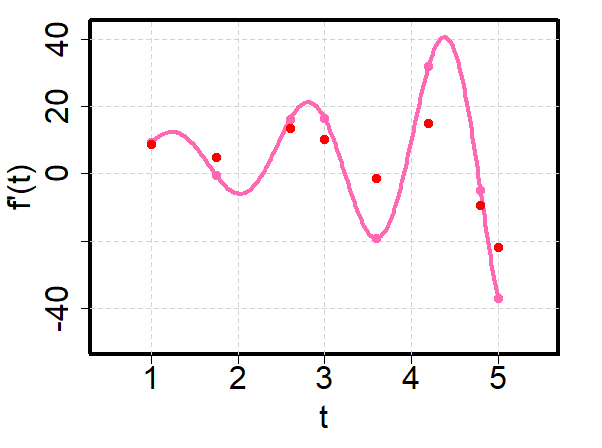}
	\caption{{\footnotesize The image on the left shows, for each point of the trajectory, a small segment of the line whose slope is given by the approximation $D_j$ to the derivative. A good approximation is one for which the red line is almost tangent to the graph. We see that the approximation is lacking especially at the fifth and sixth point, but at least every approximation carries the correct sign. The image on the right shows the graph of the derivative of the underlying function (pink) as well as its approximation $D_j$ at the observation times (red dots). We see that the approximation is farthest from the graph at the fifth and sixth point, in agreement with our previous observation.}}
	\label{fig:derivative}
\end{figure}

Recall that the first derivative $f'(t_j)$ represents the slope of the line tangent to the graph of $f$ at the point $(t_j,y_j)$. The sign and magnitude of $f'(t_j)$ contains information about the monotonicity of $f(t)$ is a neighborhood of $t_j$. A positive (resp. negative) value of $f'(t_j)$ means that $f$ is increasing (resp. decreasing) in a neighborhood of $t_j$ and $|f'(t_j)|$ tells us at what rate it is doing so. In particular, it is important to realize that if $f'$ is everywhere positive (respectively negative), then $f$ is everywhere increasing (resp. decreasing). Figure \ref{fig:derivative} illustrates what the approximation $D_j$ looks like in practice.

\begin{figure}[!ht]
	\centering
	\includegraphics[scale=0.4]{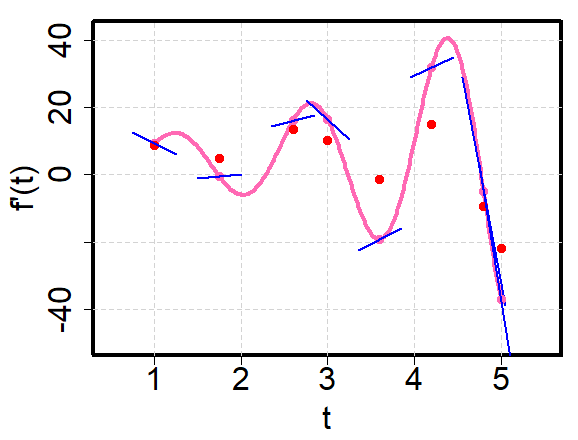}\includegraphics[scale=0.4]{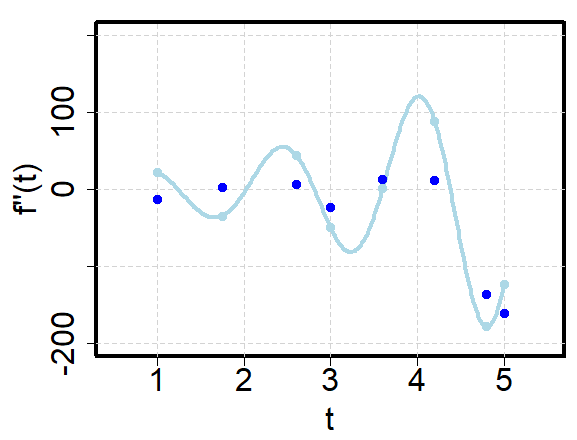}
	\caption{{\footnotesize The image on the left shows the graph of the derivative of the underlying function (pink) as well as its approximation $D_j$ at the observation times (red dots) and, in blue, a small segment of the line whose slope is approximated using $D_j^2$. A good approximation is one for which the blue line is almost tangent to the graph. The approximation is worst than that of $f'(t)$, which is to be expected since the approximation $D^2_j$ is itself built from an approximation. However, except for the first two points the signs are correct. The image on the right shows the graph of the true value of $f''(t)$ (light blue) against the approximations $D_j^2$ (blue dots).}}
	\label{fig:2ndderivative}
\end{figure}

\begin{figure}[!ht]
	\centering
	\includegraphics[scale=0.425]{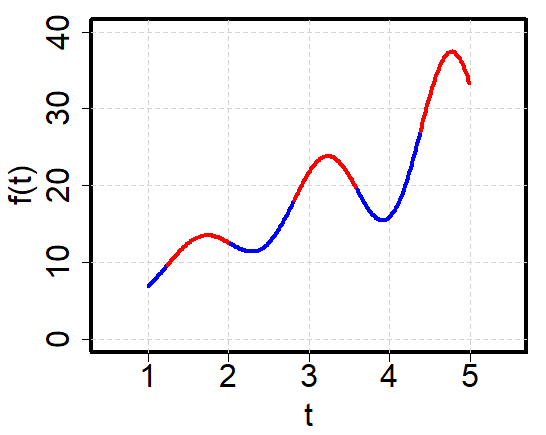}\includegraphics[scale=0.425]{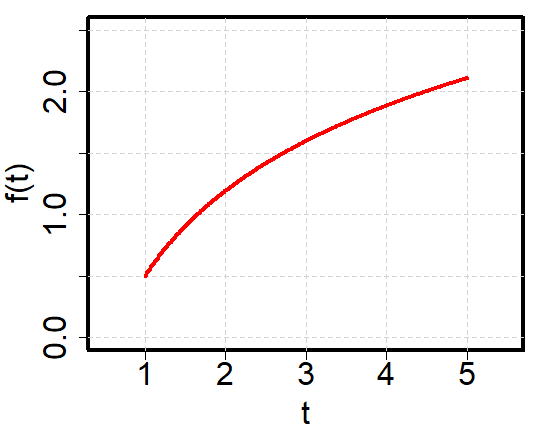}
	\caption{{\footnotesize The image on the left shows a function that has both regions of convexity (blue) and regions of concavity (red). The regions of convexity correspond exactly to the regions where the second derivative is positive while the regions of concavity are those where the second derivative is negative. The image of the right is the graph of the function $f(t)=\tfrac{1}{2}+\log t$ and is an example of a {\em globally} concave function, meaning that it has only a region of concavity. This synthetic statement is equivalent to the fact that the second derivative, $f''(t)=-1/t^2$, is everywhere negative.}}
	\label{fig:convexity}
\end{figure}

The approximation to the second derivative is illustrated in Figure \ref{fig:2ndderivative}. 
The sign of the second derivative of $f$ contains information about the shape of the graph. A positive (resp. negative) value of $f''(t_j)$ means that the graph of $f$ is convex (resp. concave) in a neighborhood of $t_j$. In particular, if $f''$ is everywhere positive (resp. negative), then $f$ is globally convex (resp. concave), as illustrated in Figure \ref{fig:convexity}. 
The magnitude of the second derivative expresses the rate of change the slope of the tangent to the graph of $f$. It is tempting but incorrect to interpret $|f''(t)|$ as a measure of the speed at which the graph of $f$ is changing directions. The notion of how fast the graph of $f$ is changing directions is best captured by the rate of change (in radian per unit time) of the orientation of the tangent line. This later quantity, the {\em curvature} of the graph, is expresible as the ratio $|f''(t)|/(1+|f'(t)|^2)^{3/2}$. Therefor, in the strictest sense, comparing the magnitude of second derivatives at different times or of different functions, only equates to a comparison of curvatures when the first derivatives coincide. 

We now proceed to describe the twenty measures that we propose to use to cluster trajectories. For each one, we start by presenting the target measure, defined at the level of the underlying function, that the trajectory measure seeks to approximate. We discuss its interpretation and provide a graphical representation when deemed helpful. Finally, we give the formula for the trajectory measure.
\begin{center}
\noindent\rule{12cm}{0.4pt}
\end{center}

\noindent $m_1$, the {\bf maximum}. The functional measure is $\max_{t\in[a,b]}f(t)$, the largest value assumed by the underlying function. The trajectory measure is
$$
m_1:=\max_{1\leq j\leq N}y_j,
$$
the largest value assumed by the trajectory. 

\begin{center}
\noindent\rule{12cm}{0.4pt}
\end{center}

\noindent $m_{2}$, the {\bf minimum}. The functional measure is $\min_{t\in[a,b]}f(t)$, the smallest value assumed by the underlying function. The trajectory measure is
$$
m_{2}:=\min_{1\leq j\leq N}y_j,
$$
the smallest value assumed by the trajectory. 

\begin{figure}[!ht]
	\centering
	\includegraphics[scale=0.4]{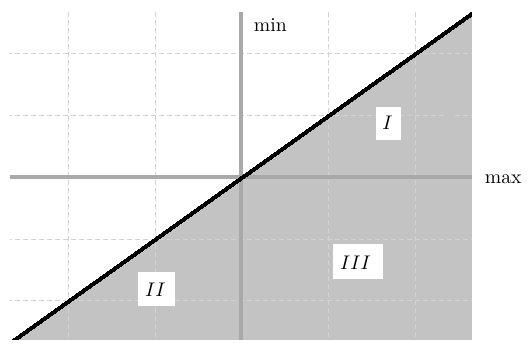}
	\caption{{\footnotesize The possible values for the point $(m_1,m_2)$. The black line suggests a constant underlying function. Region $I$suggests a positive underlying function. Region $II$ suggests a negative underlying function. In region $III$, the underlying function assumes both positive and negative values.}}
	\label{fig:3_regions_sober2}
\end{figure}

It is insightful to analyse $m_1$ and $m_2$ in tandem. In the Cartesian plane, the only allowable region for the point of coordinates $(m_1,m_2)$ is on or below the identity line. Points that fall directly on this line are points for which $m_1=m_2$ and hence they correspond to constant trajectories. Moreover, in general, $(m_1,m_{2})$ falls either in the first, third or fourth quadrant (Figure \ref{fig:3_regions_sober2}). The corresponding interpretation is that the trajectory is either positive, negative or assumes both positive and negative values.
\begin{center}
\noindent\rule{12cm}{0.4pt}
\end{center}

\noindent
$m_3$, the {\bf range}. The functional measure is the difference between the largest and the smallest values assumed by the underlying function,
$$
\rho[f]:=\max_{t\in[a,b]}f(t)-\min_{t\in[a,b]}f(t).
$$
A small (absolute) value of the range is synonymous to a function that is nearly constant. The corresponding trajectory measure is
$$
m_3:=m_1 - m_2
$$
the difference between the largest and the smallest values assumed by the trajectory.
\begin{center}
\noindent\rule{12cm}{0.4pt}
\end{center}

\begin{figure}[!ht]
	\centering
	\includegraphics[scale=0.4]{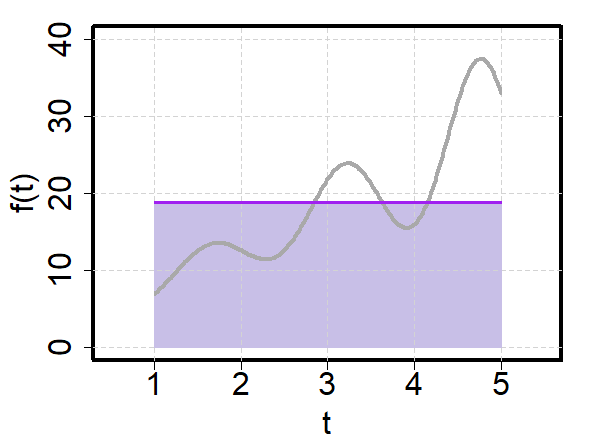}
	\caption{{\footnotesize The mean of $f$ as the height of the rectangle whose area equals the area under the curve of $f$. In the present case, $\mu[f] = 18.8$.}}
	\label{fig:mean}
\end{figure}

\noindent
$m_4$, the {\bf mean}. The functional measure is the mean value of the underlying function,
\begin{equation}\label{functional_mean}
\mu[f]:=\frac{1}{b-a}\int_a^bf(t)\,dt.
\end{equation}
A useful geometric interpretation of the means is that it is the (signed) height of the rectangle of base $[a,b]$ whose area coincides with the (signed) area under the graph of $f$ (Figure \ref{fig:mean}). Because of this, we can think of $m_3$ as a crude measure of the overall height of $f$. The trajectory measure is
    $$
    m_4:=\frac{1}{t_N-t_1}\sum_{j=1}^{N-1}\frac{y_j+y_{j+1}}{2}(t_{j+1}-t_j).
    $$
\begin{center}
\noindent\rule{12cm}{0.4pt}
\end{center}

\noindent
$m_5$, the {\bf standard deviation}. The functional measure is
\begin{equation}\label{defn_sigma}
\sigma[f]:=\left(\frac{1}{b-a}\int_a^b|f(t) - \mu[f]|^2\,dt\right)^{1/2},
\end{equation}
where $\mu[f]$ is the mean of $f$, as defined in \eqref{functional_mean}. As with the range, if $f$ is constant, the standard deviation will be small. However, the general relation between the range and the standard deviation is that, for a given value $\rho[f]$ of the range, $\sigma[f]$ can take any value in $(0,\rho[f]^2/4]$. Therefor, the standard deviation is useful in differentiating functions that share a common range. For the trajectory measure approximating the standard deviation, we have
    $$
    m_5 := \left(\frac{1}{t_N-t_1}\sum_{j=1}^{N-1}\frac{\left(y_j-m_4\right)^2 + \left(y_{j+1}-m_4\right)^2}{2}(t_{j+1}-t_j)\right)^{1/2}.
    $$
\begin{center}
\noindent\rule{12cm}{0.4pt}
\end{center}

\noindent
$m_6$, the {\bf slope of the best affine approximation}. Consider the line $\beta_0^*+\beta_1^*t$ which minimizes the $L^2$ distance to $f$. That is, the functional measure is $\beta_1^*$, as defined by
\begin{equation}\label{defn_betas}
(\beta_0^*,\beta_1^*,) = \argmin_{(\beta_0,\beta_1)}I(\beta_0,\beta_1),
\end{equation}
where 
\begin{equation}\label{defn_I}
I(\beta_0,\beta_1)=\int_a^b|\beta_0+\beta_1t - f(t)|^2\,dt.
\end{equation}
To construct a trajectory measure approximating $\beta_1^*$, we first approximate the function $I$ by
\begin{equation}\label{defn_I_hat}
\hat{I}(\beta_0,\beta_1)=\sum_{j=1}^{N-1}\frac{(\beta_0+\beta_1t_j - y_{j})^2+(\beta_0+\beta_1t_{j+1} - y_{j+1})^2}{2}(t_{j+1} - t_j).
\end{equation}
We then set $m_6$ equal to $\hat{\beta}_1^*$, as defined by
\begin{equation}\label{defn_betas_hat}
(\hat{\beta}_0^*,\hat{\beta}_1^*) = \argmin_{(\beta_0,\beta_1)}\hat{I}(\beta_0,\beta_1).
\end{equation}
An explicit formula for $m_6$ is derived in Appendix B. Figure \ref{fig:BAA2} illustrates the best affine approximation of a function and its estimation using a trajectory.
\begin{figure}[!ht]
	\centering
	\includegraphics[scale=0.425]{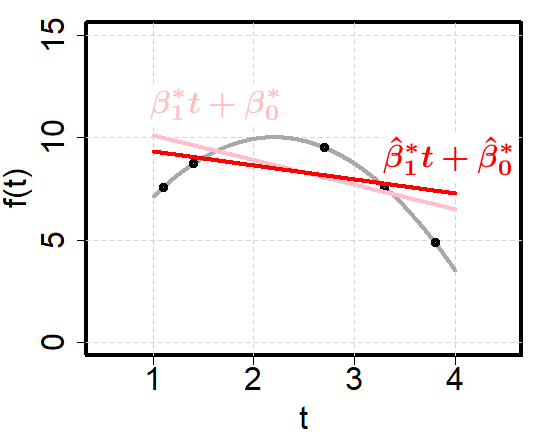}
	\caption{{\footnotesize Superimposed on the trajectory and underlying function from Figure \ref{fig:grand_example} are the best affine approximation of the underlying function (pink) and its estimation using the trajectory (red).}}
	\label{fig:BAA2}
\end{figure}
\begin{center}
\noindent\rule{12cm}{0.4pt}
\end{center}

\noindent
$m_7$, the {\bf intercept of the best affine approximation}. The functional measure is $\beta_0^*$ as defined by \eqref{defn_betas}. The trajectory measures is $\hat{\beta}_0^*$, as defined by \eqref{defn_betas_hat}. In Appendix B, it is shown that a formula for $m_7$ is
$$
m_7 = \frac{1}{t_N-t_1}\sum_{j=1}^{N-1}\frac{(y_{j}-m_6t_j) + (y_{j+1}-m_6t_{j+1})}{2}(t_{j+1}-t_j).
$$
\begin{center}
\noindent\rule{12cm}{0.4pt}
\end{center}

\begin{figure}[!ht]
	\centering
	\includegraphics[scale=0.3]{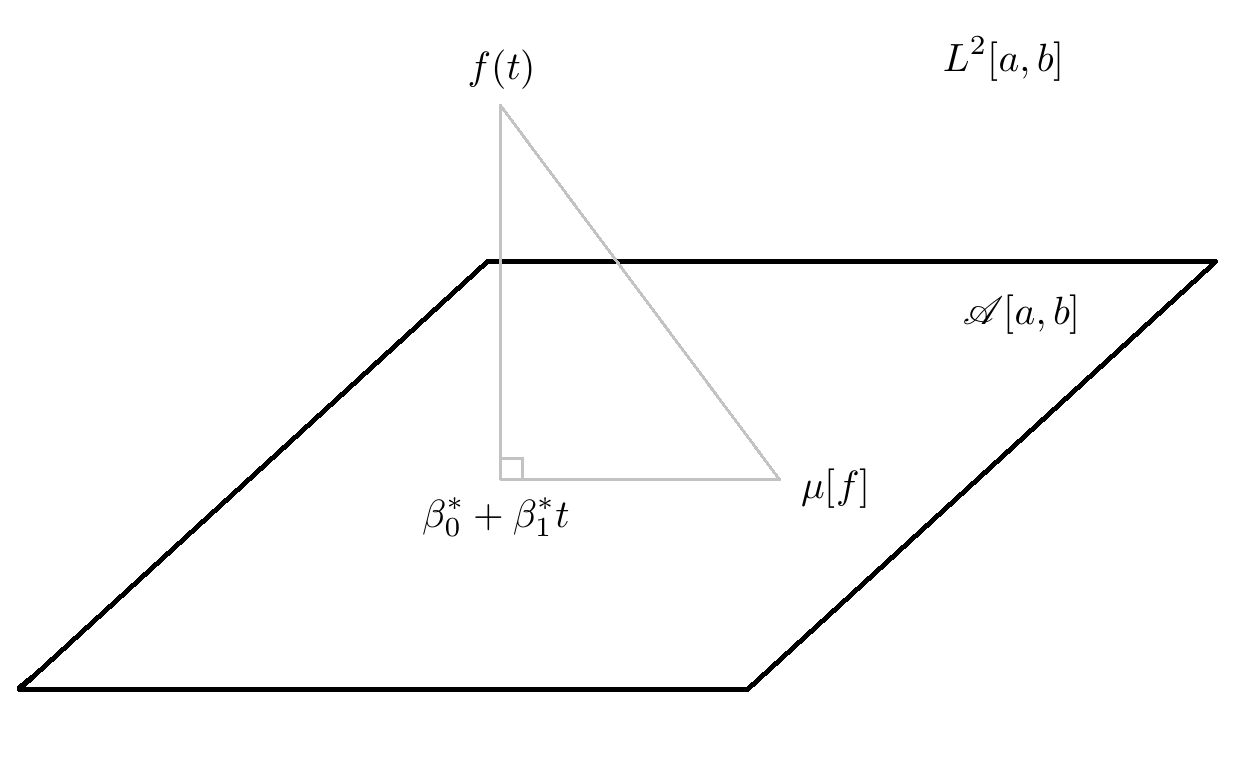}
	\caption{{\footnotesize Let $\mathscr{A}[a,b]\subset L^2[a,b]$ denote the subspace of affine functions. The affine function closest to $f(t)$ is the orthogonal projection of $f$ onto $\mathscr{A}[a,b]$. Let $f^{\perp}(t):=\beta_0^*+\beta_1^*t$ denotes said orthogonal projection. Since the constant function $\mu[f]$ belongs to $\mathscr{A}[a,b]$, the functions $f-f^{\perp}$ and $f^{\perp} - \mu[f]$ are orthogonal to each other. Therefor, we have the Pythagorean decomposition $\|f-\mu[f]\|^2 =\|f - f^{\perp}\|^2 + \|f^{\perp}-\mu[f]\|^2$ (\cite{Conway} Theorem 2.6).}}
	\label{fig:best_linear_approx}
\end{figure}

\noindent
$m_8$, the {\bf proportion of variance explained by the affine approximation}. The functional measure is
$$
R[f]^2=\frac{\int_a^b|\beta_0^*+\beta_1^*t-\mu[f]|^2\,dt}{\int_a^b|f(t)-\mu[f]|^2\,dt},
$$
where $\mu[f]$ is defined in \eqref{functional_mean} and $\beta_0^*,\beta_1^*$ are the coefficients of the affine function closests to $f$ defined by \eqref{defn_betas}. If the denominator is 0, then $f$ is constant, in which case the the numerator is also 0. In this situation, $R[f]^2$ is defined to be 1. This quantity is analogous to the coefficient of determination in linear regression. Also in analogy with the sum of squares decomposition in linear regression, we have the decomposition (see Figure \ref{fig:best_linear_approx})
\begin{equation}\label{functional_SS_dec}
\int_a^b|f(t)-\mu[f]|^2\,dt = \int_a^b|f(t) - \beta_0^*-\beta_1^*t|^2\,dt + \int_a^b|\beta_0^*+\beta_1^*t-\mu[f]|^2\,dt
\end{equation}
This identity implies that $R[f]^2$ is bounded between 0 and 1 and the interpretation that the greater $R[f]^2$ is, the closer $f$ is to its affine approximation. We approximate $R[f]^2$ by the trajectory measure
$$
m_8 := \frac{1}{m_5^2}\sum_{j=1}^{N-1}\frac{|m_7+m_6t_j-m_4|^2+|m_7+m_6t_{j+1}-m_4|^2}{2}(t_{j+1}-t_j).
$$
\begin{center}
\noindent\rule{12cm}{0.4pt}
\end{center}

\noindent
$m_{9}$, the {\bf rate of intersection with the best affine approximation}. The purpose of this functional measure is to reflect oscillatory behaviors. We define it as the ratio of the number of times that the graph of $f$ crosses the graph of the affine approximation to the lenght $b-a$ of the time interval. For example, in Figure \ref{fig:BAA2} $a=1$, $b=4$ and there are 2 intersections so the rate of intersection is 0.5. To define the corresponding trajectory measure, we assume that $f(t)$ is piecewise linear over the subintervals $[t_j,t_{j+1}]$ and count the number of crossings with the line $m_6t+m_7$. We only have to be careful with how we handle the case where some points land directly on the line. To make this rigourous, we introduce the {\em residuals} $r_j$, defined by 
\begin{equation}\label{defn_centered_traj}
r_j:=y_j-(m_7+m_6t_j).
\end{equation}
and define a crossing indicator 
$$
\chi_j=\left\{
\begin{array}{cc}
1 & \text{if $r_j\times r_{k}<0$ for $k$ the smallest index with $k>j$ and $r_k\neq 0$} \\
0 & \text{otherwise}
\end{array}
\right. .
$$
This is 1 if $r_j$ is non zero and if the next nonzero residual has sign opposite that of $r_j$ (implying that $f$ has crossed the line somewhere between these two observation times). The trajectory measure we seek is then
$$
m_{9} := \frac{1}{t_N-t_1}\sum_{j=1}^{N-1}\chi_j.
$$

\begin{center}
\noindent\rule{12cm}{0.4pt}
\end{center}

\noindent
$m_{10}$, the {\bf net variation per unit of time}. The functional measure is
$$
\frac{f(b)-f(a)}{b-a},
$$
which we approximate with
$$
m_{10}:=\frac{y_N-y_1}{t_N-t_1}.
$$
According to the fundamental theorem of calculus, 
$$
f(b)-f(a) = \int_a^bf'(t)\,dt,
$$
so $m_{10}$ can also be though of as approximating the mean derivative of $f$.
\begin{center}
\noindent\rule{12cm}{0.4pt}
\end{center}

\noindent
$m_{11}$, the {\bf contrast between the late variation and the early variation}. The concept of early and late variation requires that the underlying function comes with a distinguished (user supplied) time $t^*\in (a,b)$, the {\em mid point}, splitting the domain into the {\em early phase} $[a,t^*]$ and the {\em late phase} $[t^*,b]$. We may then consider the functional measure
$$
[f(b)-f(t^*)] - [f(t^*)-f(a)] = f(b)-2f(t^*)-f(a)
$$
contrasting the net variation occuring in the late phase with that occuring in the early phase. A positive contrast means that the net variation of $f$ is greater in the late phase than it is in the early phase and vice-versa if the contrast is negative. At the trajectory level, we must first approximate the mid point with the closest observation times; let's denote it by $t_{m}$. We then set
$$
m_{11}: = y_N-2y_{m}-y_1.
$$
\begin{center}
\noindent\rule{12cm}{0.4pt}
\end{center}

\begin{figure}[!ht]
	\centering
	\includegraphics[scale=0.425]{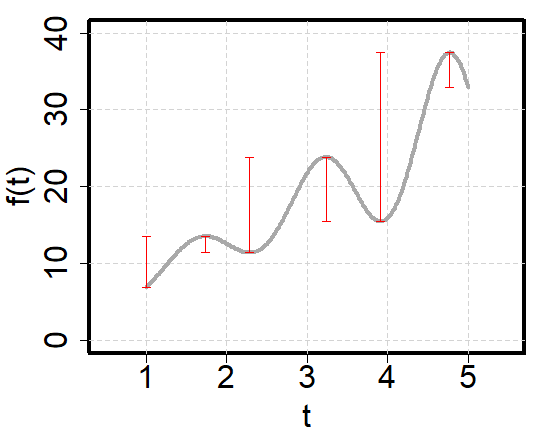}
	\caption{{\footnotesize The total variation of $f$ is the total lenght of the horizontal segments (in red). In this case, this number is approximately $6.6 + 2.1+ 12.5 + 8.4 +21.9 + 4.5=56$. }}
	\label{fig:total_variation}
\end{figure}

\noindent
$m_{12}$, the {\bf total variation per unit of time}. The total variation of $f$ is defined as the number
\begin{equation}\label{total_var_defn}
V[f]=\sup \sum_j|f(\tau_{j+1}) -f(\tau_j)|,
\end{equation}
where the supremum is over every partition $a=\tau_1<\ldots<\tau_K=b$ of $[a,b]$. Roughly speaking, $V[f]$ is the amount by which the graph of $f$ bobs up and down so a large $V[f]$ is indicative of oscillatory behavior. Figure \ref{fig:total_variation} shows an example of how $V[f]$ is computed. 
The functional measure is $V[f]/(b-a)$, which we approximate with the trajectory measure
$$
m_{12}:=\frac{1}{t_N-t_1}\sum_{j=1}^{N-1}|y_{j+1} -y_j|.
$$
Provided $+\infty$ is allowed as a value, formula \eqref{total_var_defn} is valid for any function $f$. However, in the situation which concerns us, $f$ is of class $C^2$ and, in that case, an alternative expression for $V[f]$ is $\int_a^b|f'(t)|\,dt$. Therefor, $m_{12}$ can also be thought of as approximating the mean absolute derivative of $f$.
\begin{center}
\noindent\rule{12cm}{0.4pt}
\end{center}

\begin{figure}[!ht]
	\centering
	\includegraphics[scale=0.4]{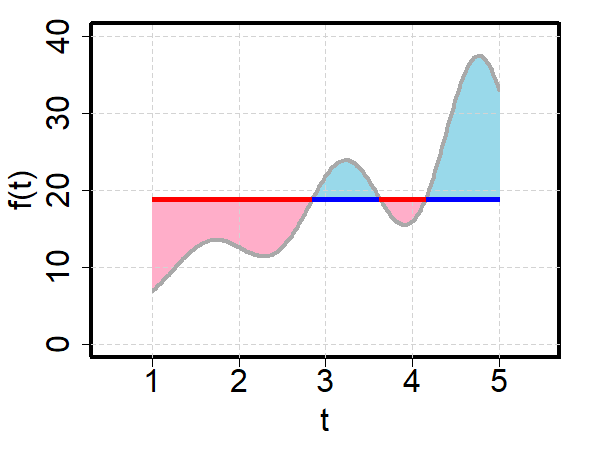}
	\caption{{\footnotesize The area under the mean (pink) is the same as the area above the mean (light blue). However, the time spent above the mean (total lenghts of the blue segments) is less than the time spent below the mean (total lenghts of the red segments). This is symptomatic of an upward spike, and indeed, there it is at the end.}}
	\label{fig:spikiness}
\end{figure}

\noindent
$m_{13}$, the {\bf spikiness}. Intuitively, a function $f$ has "spikes" if it has regions where it momentarily ventures far from its mean. A spike is "upward" (resp. "downward") if $f$ is greater (resp. lesser) than $\mu[f]$ when it spikes. One way that spikes can be detected is by comparing the amount of time that $f(t)>\mu[f]$ to the amount of time that $f(t)<\mu[f]$ (Figure \ref{fig:spikiness}). 
Indeed, almost by definition of the mean, we have the relation
$$
\frac{1}{b-a}\int_a^b(f(t)-\mu[f])\,dt = 0.
$$
This fact can be interpreted geometrically by stating that the area under the graph and above the mean is the same as the area above the graph and below the mean. When there is an upward spike, $f$ soars high above the mean, generating high amounts of area per unit time. As a result, if $f$ has more upward spikes than it has downward spikes, the time spent above the mean must be less than the time spent below the mean. Conversely, if $f$ has more downward spikes than it has upward spikes, the time spent above the mean must be greater than the time spent below the mean. Therefor, a natural candidate for detecting spikes is the difference between the relative amount of time spent above the mean and the relative amount of time spent below the mean.

\begin{figure}[!ht]
	\centering
	\includegraphics[scale=0.4]{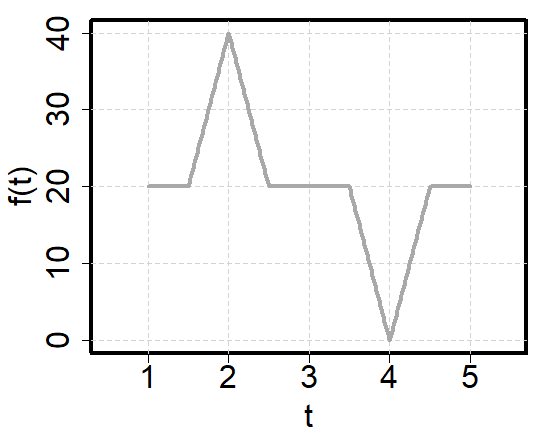}\includegraphics[scale=0.4]{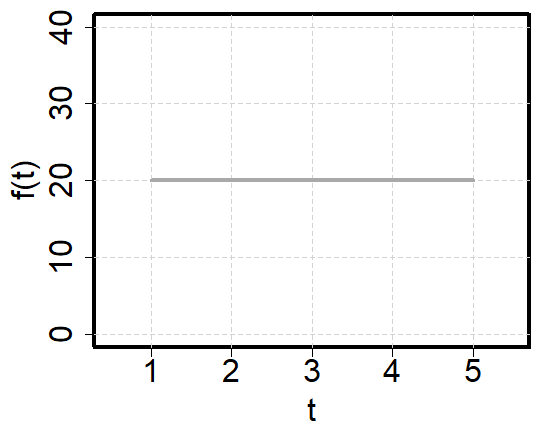}
	\caption{{\footnotesize The function on the left has an upward spike and a downward spike of equal magnitude. The function on the right has no spikes at all. Yet, $\Pi=0$ in both cases, an illustration of the fact that $\Pi$ measures the "net spikiness".}}
	\label{fig:net_spikiness}
\end{figure}

Rigorously, the time spent above the mean is the (Lebesgue) measure of the set $\mathcal{S}_+:=\{t\in [a,b] \ | \ \mu[f]<f(t)<+\infty \}$, which we denote $|\mathcal{S}_+|$. Likewise,  the time spent below the mean is $|\mathcal{S}_-|$, the measure of the set $\mathcal{S}_-:=\{t\in [a,b] \ | \ -\infty<f(t)<\mu[f] \}$. A small (resp. large) value of $|\mathcal{S}_+|$ relative to $|\mathcal{S}_-|$ is indicative of the presence of "upward (resp. downward) spikes". The functional measure of interest is
\begin{equation}\label{defn_prop_above_mean}
\Pi[f]:= \frac{|\mathcal{S}_+| - |\mathcal{S}_-|}{|\mathcal{S}_+| + |\mathcal{S}_-|} \in [-1,1].
\end{equation}
The only situation where one of $\mathcal{S}_{\pm}$ has measure 0 is when $f$ is constant, in which case both $\mathcal{S}_{\pm}$ have measure 0 and we set $\Pi[f]:=0$. We interpret a large--in absolute value--negative value of $\Pi[f]$ as indicative of the presence of "upward spikes" and we interpret a large positive value of $\Pi[f]$ as indicative of the presence of "downward spikes". One thing that is important to understand about about $\Pi[f]$ is that it really is a measure of {\em net} spikiness. Because if $f$ has both upward and downward spikes in equal measure, then $\Pi[f]=0$, the same as if $f$ had no spikes at all (Figure \ref{fig:net_spikiness}). 

\begin{figure}[!ht]
	\centering
	\includegraphics[scale=0.4]{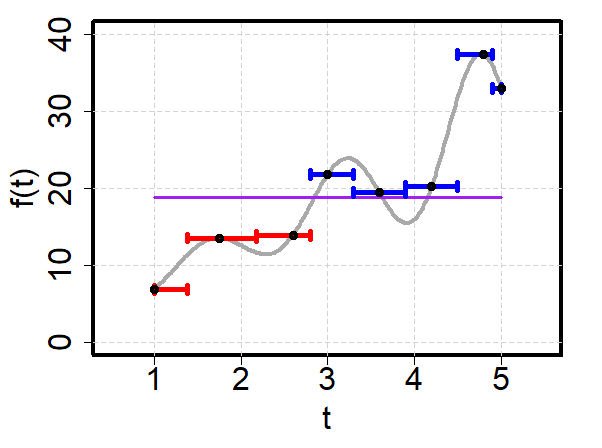}
	\caption{{\footnotesize If $f$ is below (resp. above) the mean at two subsequent observation times, we assume that the same holds at all intervening times. On the contrary, if $f$ passes from below the mean to above the mean (or vise versa), we assume that it crosses the mean precisely at the midpoint between the two observation times. In the above plot, this leads to approximating $|\mathcal{S}_+|$ as the total lenght of the blue segments and to approximating $|\mathcal{S}_-|$ as the total lenght of the red segments. If two consecutive points were to lie directly on the mean, the time interval between them would count towards neither  $|\mathcal{S}_+|$ nor  $|\mathcal{S}_-|$.}}
	\label{fig:approximating_Pi}
\end{figure}

To define a corresponding trajectory measure, we need to approximate $|\mathcal{S}_{\pm}|$. To this end, whenever $y_j$ and $y_{j+1}$ are on opposite sides of the mean, we assume that $f(t)$ crosses the mean at the mid point between $t_{j}$ and $t_{j+1}$. In case $y_j$ lands precisely on the mean, we assume that $f(t)$ has the same sign as $y_{j-1}$ on $[t_{j-1},t_j]$ and the same sign as $y_{j+1}$ on $[t_j,t_{j+1}]$. Figure \ref{fig:approximating_Pi} shows how these assumptions determine an approximation of $|\mathcal{S}_{\pm}|$.
Putting this in equations, we have
\begin{align*}
|\mathcal{S}_{+}|\approx & \ \mathbb{I}(y_1>m_4)\frac{t_2 - t_1}{2} + \sum_{j=2}^{N-1}\mathbb{I}(y_j>m_4)\frac{t_{j+1} - t_{j-1}}{2} \\
&+ \mathbb{I}(y_N>m_4)\frac{t_N - t_{N-1}}{2}
 + \mathbb{I}(y_1=m_4)\mathbb{I}(y^0_2>0)\frac{t_2-t_1}{2} \\
& + \sum_{j=2}^{N-1}\mathbb{I}(y_j=m_4)\left\{\mathbb{I}(y_{j-1}>m_4)\frac{t_j-t_{j-1}}{2} + \mathbb{I}(y_{j+1}>m_4)\frac{t_{j+1}-t_j}{2} \right\} \\
& + \mathbb{I}(y_N=m_4)\mathbb{I}(y_{N-1}>m_4)\frac{t_N-t_{N-1}}{2},
\end{align*}
\begin{align*}
|\mathcal{S}_{-}|\approx & \ \mathbb{I}(y_1<m_4)\frac{t_2 - t_1}{2} + \sum_{j=2}^{N-1}\mathbb{I}(y_j<m_4)\frac{t_{j+1} - t_{j-1}}{2} \\
&+ \mathbb{I}(y_N<m_4)\frac{t_N - t_{N-1}}{2}
 + \mathbb{I}(y_1=m_4)\mathbb{I}(y^0_2>0)\frac{t_2-t_1}{2} \\
& + \sum_{j=2}^{N-1}\mathbb{I}(y_j=m_4)\left\{\mathbb{I}(y_{j-1}<m_4)\frac{t_j-t_{j-1}}{2} + \mathbb{I}(y_{j+1}<m_4)\frac{t_{j+1}-t_j}{2} \right\} \\
& + \mathbb{I}(y_N=m_4)\mathbb{I}(y_{N-1}<m_4)\frac{t_N-t_{N-1}}{2}.
\end{align*}
Injecting these approximations into \eqref{defn_prop_above_mean} gives us $m_{13}$.

\begin{center}
\noindent\rule{12cm}{0.4pt}
\end{center}

\noindent
$m_{14}$, the {\bf maximum of the first derivative}. The functional measure is $\max_{t\in[a,b]}f'(t)$, the largest value assumed by the derivative. A large value means that $f(t)$ has regions of rapid growth. A negative (resp. non positive) value indicates that $f$ is decreasing (resp. non-increasing). Note that if the value is positive, we can only infer something about the local behavior of $f$ whereas from a non positive value we can infer something about the {\em global} behavior of $f$. The trajectory measure is
$$
m_{14}:=\max_{1\leq j\leq N}D_j.
$$
\begin{center}
\noindent\rule{12cm}{0.4pt}
\end{center}

\begin{figure}[!ht]
	\centering
	\includegraphics[scale=0.4]{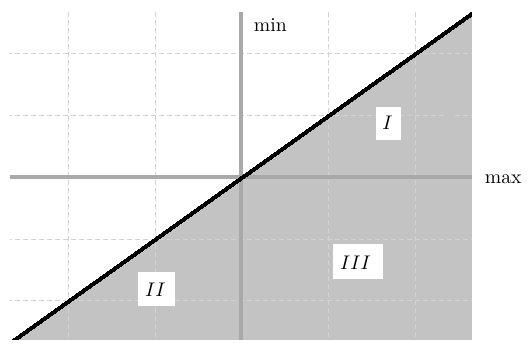}
	\caption{{\footnotesize The possible values for the point $(m_{14},m_{15})$. The black line suggests an affine underlying function. Region $I$ (resp . $II$) suggests an increasing (resp. decreasing) underlying function. In region $III$, the underlying function assumes both regions of growth and of decline.}}
	\label{fig:3_regions_sober2bis}
\end{figure}

\noindent
$m_{15}$, the {\bf minimum of the first derivative}. The functional measure is $\min_{t\in[a,b]}f'(t)$, the smallest value assumed by the derivative. A large (in absolute value) negative value means that $f(t)$ has regions of rapid decline. A positive (resp. non negative) value indicates that $f$ is increasing (resp. non-decreasing). The trajectory measure is
$$
m_{15}:=\min_{1\leq j\leq N}D_j.
$$
As with $m_1$ and $m_2$, the position of the point $(m_{14},m_{15})$ in the Cartesian plane carries information about the trajectory. The only allowable region for this point is on or below the identity line. Points that fall directly on this line are points with $m_{14}=m_{15}$, which suggests a linear trajectory. In general, $(m_{14},m_{15})$ falls either in the first, third or fourth quadrant (Figure \ref{fig:3_regions_sober2bis}). The corresponding interpretation is that the trajectory is either increasing, decreasing or has both regions where it is increasing and decreasing (and, in particular, it has local extremas). 
\begin{center}
\noindent\rule{12cm}{0.4pt}
\end{center}

\noindent
$m_{16}$, the {\bf standard deviation of the first derivative}. The functional measure is 
$$
\sigma[f']:=\left(\frac{1}{b-a}\int_a^b|f'(t) - \mu[f']|^2\,dt \right)^{1/2}.
$$
A small $\sigma[f']$ is synonymous with an $f$ that's nearly linear. In this capacity, $\sigma[f']$ is redundant with $R[f]^2$. However, $\sigma[f']$ serves another purpose, which is to discriminate between functions that have the same minimum and maximum derivative. Unlike $\min f'$ and $\max f'$ which contain information about a single point of $f$, $\sigma[f']$ is an amalgamation of information collected all along the domain. Because of this, it can distinguish functions that have the same min and max first derivative but that are otherwise very different. The trajectory measure is
$$
m_{16}:=\left( \frac{1}{t_N-t_1}\sum_{j=1}^{N-1}\frac{(D_j-\hat{\mu}[f'])^2 + (D_{j+1}-\hat{\mu}[f'])^2}{2}(t_{j+1}-t_j) \right)^{1/2},
$$
where
$$
\hat{\mu}[f'] = \frac{1}{t_N-t_1}\sum_{j=1}^{N-1}D_j(t_{j+1}-t_j).
$$
\begin{center}
\noindent\rule{12cm}{0.4pt}
\end{center}

\noindent
$m_{17}$, the {\bf derivative's net variation per unit of time}. It can be useful to have a way of encoding the direction of the graph of $f$ at the end points of its domain. We consider the functional measure 
$$
\frac{f'(b)-f'(a)}{b-a}
$$
which we approximate with
$$
m_{17}:=\frac{D_N-D_1}{t_N-t_1}.
$$
Typically, a large positive (resp. negative) value of $m_{17}$ signifies that $f$ is decreasing (resp. increasing) at $t=t_1$ but increasing (resp. decreasing) at $t=t_N$. A small value of $m_{17}$ means that $f$ is either increasing or decreasing at both at $t=t_1$ and $t_N$.

Because of the fundamental theorem of calculus, we have
$$
f'(b)-f'(a) = \int_a^bf''(t)\,dt,
$$
so that $m_{17}$ can also be though of as approximating the mean second derivative of $f$.

\begin{center}
\noindent\rule{12cm}{0.4pt}
\end{center}

\noindent
$m_{18}$, the {\bf maximum of the second derivative.} The functional measure is $\max_{t\in[a,b]}f''(t)$, the largest value assumed by the second derivative. A positive value means that $f(t)$ has regions of convexity. On the other hand, a negative value implies that $f''(t)$ is negative everywhere and hence that $f$ is {\em globally concave}. The trajectory measure is
$$
m_{18}:=\max_{1\leq j\leq N}D^2_j.
$$
\begin{center}
\noindent\rule{12cm}{0.4pt}
\end{center}

\begin{figure}[!ht]
	\centering
	\includegraphics[scale=0.4]{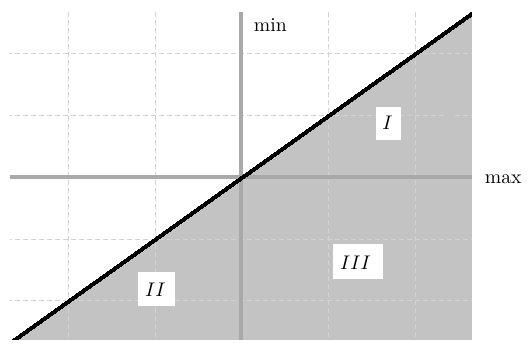}
	\caption{{\footnotesize The possible values for the point $(m_{18},m_{19})$. The identity line (in black) suggests a quadratic underlying function. Region $I$ (resp . $II$) suggests a globally convex (resp. globally concave) underlying function. In region $III$, the underlying function assumes both regions of convexity and of concavity.}}
	\label{fig:3_regions_sober2bisbis}
\end{figure}

\noindent
$m_{19}$, the {\bf minimum of the second derivative.} The functional measure is $\min_{t\in[a,b]}f''(t)$, the smallest value assumed by the second derivative. A negative value means that $f(t)$ has regions of concavity. On the other hand, a positive value implies that $f''(t)$ is positive everywhere and hence that $f$ is {\em globally convex}. The trajectory measure is
$$
m_{19}:=\min_{1\leq j\leq N}D^2_j.
$$
Again, we will find it useful to interpret $m_{18}$ and $m_{19}$ simulteneously by looking at the point of coordinates $(m_{18},m_{19})$ in the Cartesian plane. The only allowable region for this point is on or below the identity line. Points that fall directly on this line are points for which $m_{18}=m_{19}$ and hence they suggest a quadratic trajectory. That is to say, they suggest that there are constants $\alpha,\beta,\gamma$ such that, for all $j$, we have $y_j =\alpha t_j^2 + \beta t_j + \gamma$. In general, $(m_{18},m_{19})$ falls either in the first, third or fourth quadrant (Figure \ref{fig:3_regions_sober2bisbis}). The corresponding interpretation is that the trajectory is either convex, concave or has both regions where it is convex and concave and, in particular, it has inflexion points.
\begin{center}
\noindent\rule{12cm}{0.4pt}
\end{center}

\noindent
$m_{20}$, the {\bf standard deviation of the second derivative}. The functional measure is 
$$
\sigma[f'']:=\left(\frac{1}{b-a}\int_a^b|f''(t) - \mu[f'']|^2\,dt \right)^{1/2}.
$$
A small $\sigma[f'']$ is synonymous with an $f$ that's nearly quadratic. A nonzero $\sigma[f'']$ can mean many things for $f$ but nevertheless, $\sigma[f'']$ can be useful at discriminate between functions that have similar minimum and maximum second derivative. The trajectory measure is
$$
m_{20}:=\left( \frac{1}{t_N-t_1}\sum_{j=1}^{N-1}\frac{(D^2_j-\hat{\mu}[f'])^2 + (D^2_{j+1}-\hat{\mu}[f'])^2}{2}(t_{j+1}-t_j) \right)^{1/2},
$$
where
$$
\hat{\mu}[f''] = \frac{1}{t_N-t_1}\sum_{j=1}^{N-1}D^2_j(t_{j+1}-t_j).
$$

\section{A clustering algorithm for the measures}\label{section_clustering}

Consider a set of trajectories $\bm{y}_i\in \mathbb{R}^{N_i}$, $i=1,\ldots,n$ that we wish to partition into clusters. Assume that, for each trajectory, we have computed a corresponding set of measures $\bm{m}_i=(m_{\ell_1},\ldots, m_{\ell_d})$, $1\leq\ell_1<\ldots<\ell_d\leq20$ among the measures defined in section \ref{section_measures}. At this point, it remains only to find clusters in the space of measures. We encourage users who are so inclined to experiment with various clustering techniques to find the one that is most aligned with their goal and their data. That said, among the myriads of clustering algorithms available \cite{Xu2015ClusteringSurvey, EstivillCastro2002WhySM}, we recommend focusing on those that are apt at identifying non convex clusters, such as DBSCAN \cite{Ester1996DBSCAN}, HDBSCAN \cite{Campello2015HDBSCAN} and Spectral Clustering \cite{Ng}. We make this recommendation based on the empirical observation that the measures often arrange themselves in non convex clusters.

We now present a version of the Spectral Clustering algorithm that requires no parameter tuning and that, we have found, works well on a variety of data sets. As a first step, it is required that a value $S_{ij}\geq 0$, representing the {\em similarity} between $\bm{m}_i$ and $\bm{m}_j$ be chosen such that $S_{ij} = S_{ji}$. The interpretation is that the greater the value of $S_{ij}$, the more $\bm{m}_i$ and $\bm{m}_j$ are considered similar.\footnote{An exception to this interpretation is the {\em self-similarities} $S_{ii}$, which do not play a role in what follows and can be set 0.} Given this, the Spectral Clustering algorithm of Ng et al. \cite{Ng} proceeds as follow (cf. \cite{Meila}).
\begin{itemize}
\item[1.] Starting from the matrix $S$, form the matrix $P$ obtained from $S$ by normalisation of its rows. That is
$
P_{ij} = S_{ij}\slash\sum_kS_{ik};
$
\item[2.] Compute the $K$ largest eigenvalues $1=\lambda_1\geq \ldots\geq \lambda_K$ of $P$ with associated orthogonal eigenvectors $\bm{u}_1,\ldots,\bm{u}_K$;
\item[3.] Construct the $n\times (K-1)$ matrix $X$ whose $k$-th column is $\bm{u}_{k+1}$;
\item[4.] Construct the mapping $\varphi$ associating $\bm{m}_i$ to the normalized $i$-th row of $X$;
\item[5.] Run the $K$-means clustering algorithm (see Appendix \ref{app_C}) on the points $\varphi(\bm{m}_i)\in \mathbb{R}^{K-1}$ to form $K$ clusters.
\end{itemize}
We now describe how to compute the similarities that we advocate for. Firstly, the  $\bm{m}_i$ are standardized. By an abuse of notation, the standardized measures are denoted by $\bm{m}_i$ still. For an integer $p$ whose value is specified below, let $S_{ij}$ equal...
\begin{itemize}
\item[i.] 1 if $\bm{m}_i$ is among the $p$ nearest neighbors of $\bm{m}_j$ {\em and} $\bm{m}_j$ is among the $p$ nearest neighbors of $\bm{m}_i$;
\item[ii.] $\tfrac{1}{2}$ if $\bm{m}_i$ is among the $p$ nearest neighbors of $\bm{m}_j$ but $\bm{m}_j$ is {\em not} among the $p$ nearest neighbors of $\bm{m}_i$;
\item[iii.] $\tfrac{1}{2}$ if $\bm{m}_j$ is among the $p$ nearest neighbors of $\bm{m}_i$ but $\bm{m}_i$ is {\em not} among the $p$ nearest neighbors of $\bm{m}_j$;
\item[iv.] 0 if neither $\bm{m}_i$ is among the $p$ nearest neighbors of $\bm{m}_j$  nor $\bm{m}_j$ is among the $p$ nearest neighbors of $\bm{m}_i$.
\end{itemize}
For the number $p$ of neighbors to consider, we recommend using
\begin{equation}\label{defn_p}
p:=
\left\{
\begin{array}{cc}
2 & \text{if $n/K<3.5$} \\
3 & \text{if $3.5\leq n/K<4.5$} \\
\max\left(4,\min\left(8,\lfloor n/2K \rfloor\right)\right) & \text{otherwise}
\end{array}
\right. .
\end{equation}
In defining $p$ this way, we try to keep $p$ at around half of the typical cluster size $n/K$ while being no greater than 8 and no smaller than 4, {\em unless} the typical cluster size is less than 4.5. In that case, $p$ is allowed to take on the values 2 and 3. Table \ref{tbl1} displays the value of $p$ for various combinations of $n$ and $K$. 

\begin{table}[h!]
\centering
\begin{tabular}{c|ccccccccc}
$K \backslash n$ & 20 & 30 & 40 & 50 & 60 & 70 & 80 & 90 & 100 \\
\hline
3  & 4 & 5 & 6 & 8 & 8 & 8 & 8 & 8 & 8 \\
4  & 4 & 4 & 5 & 6 & 7 & 8 & 8 & 8 & 8 \\
5  & 3 & 4 & 4 & 5 & 6 & 7 & 8 & 8 & 8 \\
6  & 2 & 4 & 4 & 4 & 5 & 5 & 6 & 7 & 8 \\
7  & 2 & 3 & 4 & 4 & 4 & 5 & 5 & 6 & 7 \\
8  & 2 & 3 & 4 & 4 & 4 & 4 & 5 & 5 & 6 \\
9  & 2 & 2 & 3 & 4 & 4 & 4 & 4 & 5 & 5 \\
10 & 2 & 2 & 3 & 4 & 4 & 4 & 4 & 4 & 5 \\
\end{tabular}
\caption{Values of $p$ corresponding to various combinations of the sample size $n$ and the number of groups $K$}
\label{tbl1}
\end{table}

We must warn that there are special situations, when the true cluster sizes are unbalanced, for which the choice of $p$ \eqref{defn_p} is inadequate. This occurs when $n$ is sufficiently large relative to $K$ that $n/2K\geq 8$ while at the same time having true clusters of size less than 8. In such a scenario, $p=8$, so the algorithm may fail to isolate the smaller clusters.

\subsubsection{Scale invariance}
By inspecting the definition of the measures in section \ref{section_measures}, the reader can verify that a rescaling in the units of $t$ or $y$ results in a rescaling of the measures. This means that for every measure $m$ there is a function $A(\alpha_1,\alpha_2)$ such that
$$
m[\alpha_1\bm{y}, \alpha_2\bm{t}] = A(\alpha_1,\alpha_2)m[\bm{y},\bm{t}].
$$
Since the standardization procedure is invariant under a transformation of this form, it follows that the FBTC approach as a whole is invariant under a rescaling of the units on which the trajectories and/or time are measured.

\subsubsection{Data preprocessing}
Most of the trajectory measures defined in section \ref{section_measures} are what we might call translation invariant. To be precise, let us say that a measure $m[\bm{y}, \bm{t}]$ is {\em vertically invariant} (resp. {\em horizontally invariant}) if $m[\bm{y}+c\bm{1},\bm{t}] = m[\bm{y},\bm{t}]$ (resp. $m[\bm{y},\bm{t}+c\bm{1}] = m[\bm{y},\bm{t}]$) for any constant $c$ and $\bm{1} = (1,\ldots,1)$. The measures that are not vertically invariant are the maximum ($m_1$), the minimum ($m_2$), the mean ($m_4$) and the intercept of the affine approximation ($m_7$). The only one that is not horizontally invariant is the intercept of the linear approximation ($m_7$). The implication of this is that, depending on context, if one is purely interested in clustering trajectories based on their general behavior and not on their general position along the vertical axis, it is recommended to either exclude $m_1,m_2,m_4,m_7$ from the clustering step or to work with the mean-centered trajectories $\bm{y}^0_i = \bm{y}_i-m_3[\bm{y}_i,\bm{t}_i]\bm{1}$ instead. Likewise, if one is not interested in differentiating trajectories that differ by a horizontal translation, one should either exclude $m_7$ from the clustering step or work with the horizontally shifted times $\bm{t}^0_i = \bm{t}_i-\min_j t_{1j}\bm{1}$.

\subsubsection{Outlier control}
The method is quite robust to outliers, but not entirely. This is because if a trajectory $\bm{y}_{i'}$ is an outlier, there will typically be multiple measures among $m_{1 i'},\ldots,m_{20 i'}$ which are outliers. Each such outlier $m_{\ell i'}$ will inflate the corresponding standard deviation $\mathrm{sd}(m_{\ell})$ so all the standardized measures will find themselves compressed along the $\ell$-th dimension. If the compression is sufficiently severe, the relative position of the $\bm{m}_{i}$ can be disturbed. Because of this, it is advisable to either prune out the outlier trajectories before clustering, or at least apply a capping procedure on the measures to limit the impact of outliers. A possible technique for identifying outliers is to run $K$-means on the standardized measures for a relatively large value of $K$ because outliers tend to end up in one-point clusters.


\subsubsection{Handling noise}
We have stated at the begining of section \ref{section_measures} that the trajectories are always assumed to be faithful representations of the underlying functions, in the sense that $y_j=f(t_j)$ with no measurement error or noise. If there is reason to suspect that noise is present, so that $y_j=f(t_j) + \epsilon_{ij}$, it can be worthwhile to denoise the data prior to running the algorithm. Otherwise, the algorithm might latch on to features of the data which are proper to the noise rather than the underlying functions. Or, the noise might obsure certain features of the underlying functions. However, it is also possible that the features characterizing the the clusters are still discernable through the noise. So a small amount of noise is unlikely to throw off the algorithm. Depending on the situation, denoising could be achieved for instance using wavelets or smoothing splines.

%
%

%
%
%

\subsubsection{Fuzzy FBTC}
The $K$-means algorithm described in Appendix \ref{app_C} belongs to a class of clustering algorithms commonly known as {\em hard}. This refers to the fact that the output assigns unambiguously each point to a unique cluster. By contrast, a soft clustering algorithm allows for more nuance by recognizing that the class belonging of certain points is more ambiguous than that of others. The output is, for each individual, a probability distribution representing the credence or degree of belonging to each cluster. One instance of a soft clustering algorithm is Fuzzy $K$-means, described in some details in Appendix \ref{app_D}. Now, recall that the fifth and final of the Spectral Clustering algorithm as described in section \ref{section_clustering} is to apply $K$-means on the transformed data. If Fuzzy $K$-means is applied instead, we obtain a soft version of FBTC.

\section{Examples}\label{section_examples}
We present four examples of application of FBTC. The first example consists of simulated data of our own design and intendent specifically to highlight the kind of data on which FBTC shines relative to other classification methods. The other three examples are taken from the UCR Time Series Classification Archive \cite{UCR}. The trajectories in this archive come with a group label, allowing for the comparison of the solution put forth by FBTC with a referential "true" grouping.

\subsubsection{Simutated data}
\begin{figure}[!ht]
	\centering
	\includegraphics[scale=0.4]{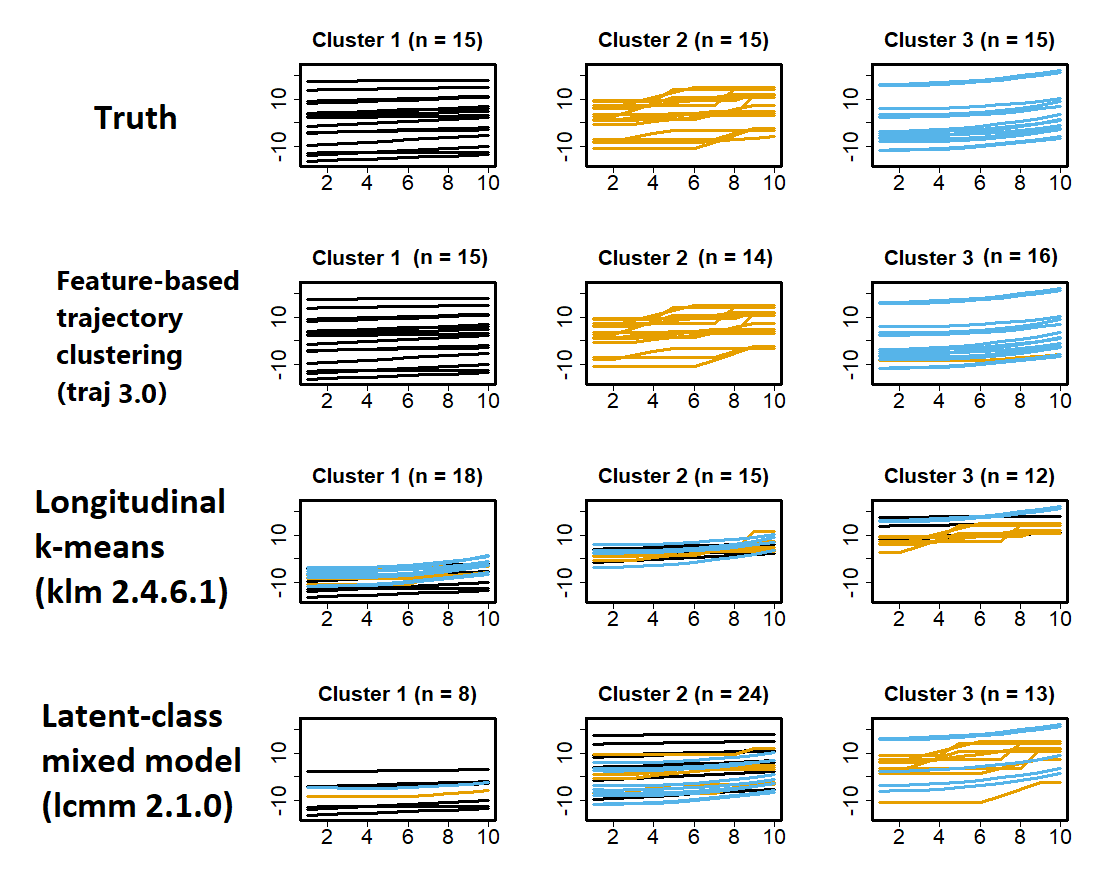}
	\caption{{\footnotesize Clusterings proposed by FBTC, klm and lcmm}}
	\label{fig:ex_sim}
\end{figure}

The simulated data set of 45 trajectories over 10 equidistant observation times and split evenly over three groups. All three groups feature monotonous trajectories with the first group being characterized by constant linear growth, the second by a stepwise increase and the third by a slow quadratic growth. Just to highlight the effect of the difference in which FBTC approaches the problem, we include the clustering solution by two popular trajectory clustering methods, $K$-means (implemented in R by {\em kml} 2.4.6.1) and a latent class mixed model (LCMM) with quadratic time dependence (implemented by {\em lcmm} 2.1.0). As can be seen in Figure \ref{fig:ex_sim}, FBTC makes a single classification mistake, assigning a curve from group 2 to group 3. By comparison, {\em kml} ignores the characterizing shapes of the groups and instead focuses on forming groups of nearby curves. As for {\em lcmm}, it is not clear to the naked eye what informed its choice of classification.

\subsubsection{Symbols}
\begin{figure}[!ht]
	\centering
	\includegraphics[scale=0.4]{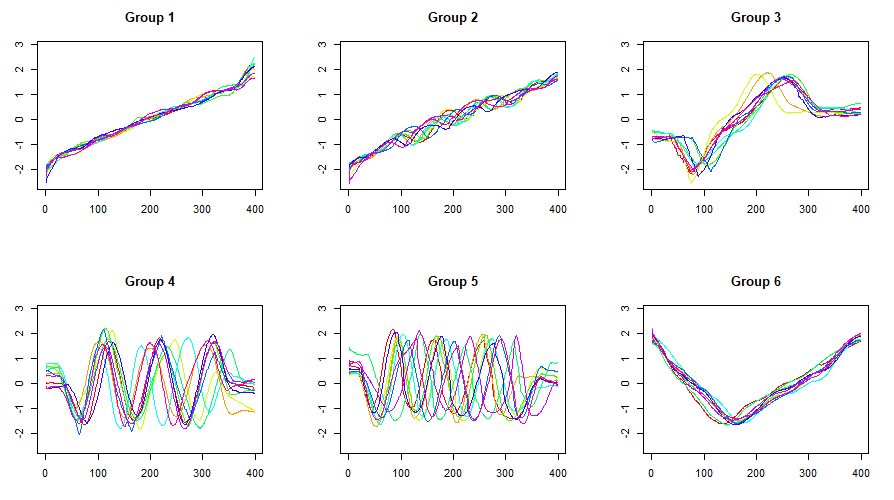}
	\caption{{\footnotesize A random sample of 10 trajectories from each of the 6 groups of the Symbols data set}}
	\label{fig:Symbols}
\end{figure}

The Symbols data set has six groups of respective sizes 173, 157, 164, 178, 159, 164. Each trajectory has 389 equidistant observations. Ten randomly chosen trajectories from each groups are illustrated in Figure \ref{fig:Symbols} with the aim of elucidating the nature of the groups. 
The clustering solution proposed by FBTC is illustrated in Table \ref{tbl1}. In this table,the $i$-th row shows how the trajectories in the $i$-th true group distributes themselves across the six groups identified by FBTC. We see that, up to small errors, FBTC recreates groups 3, 4, 5, 6 rather faithfully. However, its classificaiton of the trajectories in groups 1 and 2 does not coincide with the truth. Rather, FBTC take the trajectories in group 1 and 2 and forms two groups, of size 204 and 126, with the smaller groups being characterized by large values of $m_4$ and $m_{14}$ and small values of $m_{17}$, $m_{19}$. 

\begin{table}[]
\centering
\begin{tabular}{l|llllll}
    & 204 & 171 & 168 & 163 & 163 & 126 \\ \hline
173 & 109 &     &     &     &     & 63  \\
157 & 95  &     &     &     &     & 62  \\
164 &     &     & 1   &     & 163 &     \\
178 &     & 168 & 9   &     &     & 1   \\
159 &     & 3   & 156 &     &     &     \\
164 &     &     & 2   & 162 &     &    
\end{tabular}
\caption{The rows correspond to the true groupings whereas the columns correspond to the clustering proposed by FBTC}
\label{tbl1}
\end{table}

\subsubsection{Fungi}
\begin{figure}[!ht]
	\centering
	\includegraphics[scale=0.4]{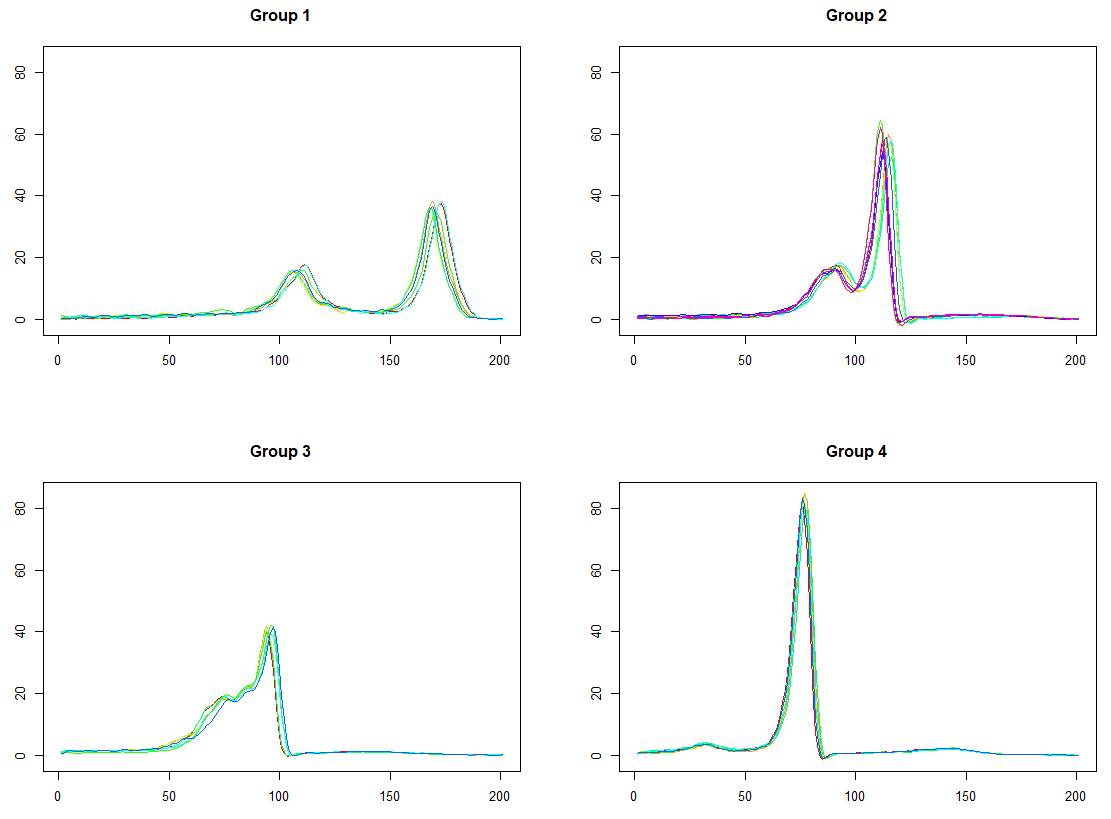}
	\caption{{\footnotesize A random sample of up to 10 trajectories from each of the 4 groups comprising the Fungi data set}}
	\label{fig:Fungi}
\end{figure}

Among the large Fungi data set, we have selected 4 groups of respective sizes 7,19,7,7. Each trajectory has 201 observations. Figure \ref{fig:Fungi} illustrate the groups. In this case, the clustering proposed by FBTC coincides perfectly with the true clustering.

\subsubsection{ShapesAll}
\begin{figure}[!ht]
	\centering
	\includegraphics[scale=0.4]{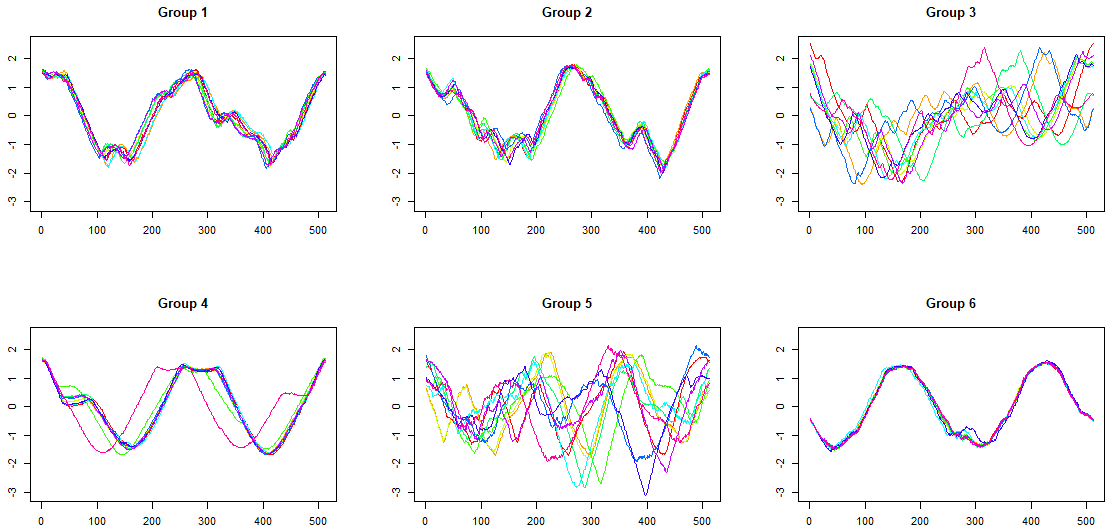}
	\caption{{\footnotesize A random sample of 10 trajectories from each of the 6 groups comprising the ShapesAll data set}}
	\label{fig:ShapesAll}
\end{figure}

The ShapesAll data set has 60 groups, each with 512 equidistant observations. We consider the six groups of size 20 labeled 15,16,17,18,19,20. They are illustrated in Figure \ref{fig:ShapesAll} where 10 random trajectories from each groups are plotted. The only difference between the clustering identified by FBTC and the trye clustering is that 2 trajectories from group 3 are assigned into group 5.

\section{Conclusion}
This paper is the first in a series of 3 companions papers about FBTC. The second paper is expected to appear by June 2026. In it, we compare FBTC to existing methods of trajectory clustering in a variety of settings through simulations. The third paper is expected to appear by October 2026 and concerns the R package traj, which implement FBTC alongside various tools facilitatings the visualisation and interpretation of the results. The R package traj also includes an option to form clusters of individuals to which are associated more than one trajectory.

\section*{Appendix A: Approximating the first and second derivatives}\label{app_A}
If a functional measure involves the first derivative $f'(t)$, we approximate it at $t=t_j$ with
$$
D_j^{\pm}:=\frac{y_{j\pm 1}-y_{j}}{t_{j\pm 1} - t_{j}} \approx \lim_{\Delta t\rightarrow 0^{\pm}}\frac{f(t_j+\Delta t) - f(t_j)}{\Delta t}
$$
approximate the right and left derivatives respectively and let
$$
w_j^{\pm}:=\frac{|t_j-t_{j\pm 1}|}{t_{j+1}-t_{j-1}}
$$
be weights reflecting the relative proximity of $t_{j\pm 1}$ to $t_j$. We approximate $f'(t_j)$ with 
\begin{equation}\label{defn_Df}
D_j := 
\left\{
    \begin{array}{cc}
    D_j^+ & \text{if $j=1$} \\
    w_j^-D^-_j + w_j^+D^+_j & \text{if $1<j<N$} \\
    D_j^- & \text{if $j=N$}
    \end{array}
    \right. .
\end{equation}

If a functional measure involves the second derivative $f''(t)$, we approximate it at $t=t_j$ using $D_j$ in the same way that we approximated $f'(t)$ using $y_j$. We set 
$$
(D^2)_j^{\pm}:=\frac{D_{j\pm 1}-D_{j}}{t_{j\pm 1} - t_{j}} \approx \lim_{\Delta t\rightarrow 0^{\pm}}\frac{f'(t_j+\Delta t) - f'(t_j)}{\Delta t}
$$
and
\begin{equation}\label{defn_D2f}
D^2_j := 
\left\{
    \begin{array}{cc}
    (D^2)_j^+ & \text{if $j=1$} \\
    w_j^-(D^2)^-_j + w_j^+(D^2)^+_j & \text{if $1<j<N$} \\
    (D^2)_j^- & \text{if $j=N$}
    \end{array}
    \right. .
\end{equation}

\section*{Appendix B: The best affine approximation}\label{app_B}
We need to solve
$$
(\hat{\beta}_0^*,\hat{\beta}_1^*) = \argmin_{(\beta_0,\beta_1)}\hat{I}(\beta_0,\beta_1),
$$
where $\hat{I}$ is defined in \eqref{defn_I_hat}.
Alternatively, this coincides with the solution of the system
$$
\frac{\partial \hat{I}(\beta_0,\beta_1)}{\partial\beta_0}=0,\quad\frac{\partial \hat{I}(\beta_0,\beta_1)}{\partial\beta_1}=0.
$$
First, we compute
\begin{align*}
\frac{\partial \hat{I}(\beta_0,\beta_1)}{\partial\beta_0} &= \sum_{j=1}^{N-1}[2\beta_0+\beta_1(t_j+t_{j+1}) - (y_j+y_{j+1}))](t_{j+1}-t_j) \\
& = 2\left[\beta_0(t_N-t_1) + \sum_{j=1}^{N-1}(\beta_1\overline{t}_j-\overline{y}_j)\Delta t_j\right],
\end{align*}
where we have introduced the notation
$$
\Delta t_j = t_{j+1}-t_j, \quad \overline{t}_j = \frac{t_j+t_{j+1}}{2},\quad \overline{y}_j=\frac{y_j+y_{j+1}}{2}.
$$
From this, we deduce that $\partial \hat{I}/\partial\beta_0=0$ if and only if 
\begin{equation}\label{beta0_as_fct_of_beta1}
\beta_0 = \frac{1}{t_N-t_1}\sum_{j=1}^{N-1}(\overline{y}_j-\beta_1\overline{t}_j)\Delta t_j.
\end{equation}
Next, we compute
\begin{align}\label{der_wrt_beta1}
\frac{\partial \hat{I}(\beta_0,\beta_1)}{\partial\beta_1} & = \sum_{j=1}^{N-1}[t_{j}(\beta_0+\beta_1t_j - y_j)+ t_{j+1}(\beta_0+\beta_1t_{j+1} - y_{j+1})](t_{j+1}-t_j) \\
& = \sum_{j=1}^{N-1}[\beta_0(t_{j}+t_{j+1}) + \beta_1(t_{j+1}^2+t_{j}^2) - (t_{j}y_{j}+t_{j+1}y_{j+1})](t_{j+1}-t_j) \\
& = 2\sum_{j=1}^{N-1}[\beta_0\overline{t}_j + \beta_1\overline{t^2}_j  - \overline{ty}_j)]\Delta t_j,
\end{align}
where we have further introduced the notation
$$
\overline{t^2}_j = \frac{t_j^2+t_{j+1}^2}{2} ,\quad \overline{ty}_j  =\frac{t_jy_j+t_{j+1}y_{j+1}}{2}.
$$
Assuming that \eqref{beta0_as_fct_of_beta1} holds, we may express $\beta_0$ in terms of $\beta_1$ in \eqref{der_wrt_beta1} and solve the equation $\partial \hat{I}/\partial\beta_1=0$ for $\beta_1$, yielding
$$
\hat{\beta}_1^* = \frac{\sum_{j=1}^{N-1}\overline{ty}_j\Delta t_j - \frac{1}{t_N-t_1}\left(\sum_{j=1}^{N-1}\overline{t}_j\Delta t_j\right)\left(\sum_{k=1}^{N-1}\overline{y}_k\Delta t_k\right)}{\sum_{j=1}^{N-1}\overline{t^2}_j\Delta t_j - \frac{1}{t_N-t_1}\left(\sum_{j=1}^{N-1}\overline{t}_j\Delta t_j\right)^2}.
$$
Injecting this back into \eqref{beta0_as_fct_of_beta1} gives
$$
\hat{\beta}_0^* = \frac{1}{t_N-t_1}\sum_{j=1}^{N-1}(\overline{y}_j-\hat{\beta}_1^*\overline{t}_j)\Delta t_j.
$$
The best affine approximation is related to ordinary least square (OLS) regression as follows. Denote $S$ the objective function for OLS,
$$
S(\beta_0,\beta_1):=\sum_{j=1}^N(\beta_0+\beta_1t_j-y_j)^2.
$$
Assuming that $\Delta t_j=\Delta t$ for all $j$, we have
$$
S(\beta_0,\beta_1)\cdot\Delta t = \sum_{j=1}^{N-1} (\beta_0+\beta_1t_j-y_j)^2\Delta t + (\beta_0+\beta_1t_N-y_N)^2\Delta t 
$$
The leftmost term on the right hand side of this equation is the left Riemann sum for the objective function $I$. Therefor, it converges to $I(\beta_0,\beta_1)$ as $\Delta t\rightarrow 0$. As for the  rightmost term, it converges to 0 as $\Delta t\rightarrow 0$. Therefor, 
\begin{equation}\label{asymptotic}
S(\beta_0,\beta_1)\cdot\Delta t - I(\beta_0,\beta_1) \rightarrow 0 \quad \text{as $\Delta t\rightarrow 0$}.
\end{equation}
This shows that, in the asymptotic limit of equally spaced points, OLS and the best affine approximation coincide. Indeed, \eqref{asymptotic} implies that if $N$ is large enough, $S$ and $I$ are almost proportional and so they have almost the same minimizer.

\section*{Appendix C: $K$-means}\label{app_C}
Consider the task of partitioning into $K$ clusters a set of points $\bm{x}_i\in\mathbb{R}^N$ living in some Euclidean space. A clustering $\mathcal{C}_1,\ldots,\mathcal{C}_K$ is considered good if the points within a given clusters are all close to one another. Its quality can be quantified by the {\em within cluster sum of squares} (WCSS)
$$
WCSS=\sum_{k=1}^K\sum_{\bm{x}\in \mathcal{C}_k}\|\bm{x}-\bm{\mu}_k\|^2,
$$
where $\bm{\mu}_k$ is the mean of $\mathcal{C}_k$. The smaller the WCSS, the better the clustering. Indeed, minimizing $WCSS$ means finding the clustering whose points are closest to their respective cluster's centers. Underlying this approach is the transitivity of proximity: by virtue of all being close to a common center, the elements of a cluster are close to one another. Unfortunately, finding the partition that minimizes WCSS by trying every possible combination is too time consuming even for data sets of modest size. A {\em $K$-means algorithm} is an algorithm that takes as input an initial clustering and applies a series of modifications that decrease the WCSS. Examples include the algorithms of MacQueen \cite{MacQueen} and that of Hartigan and Wong \cite{Hartigan_Wong}. Since the output of a $K$-means algorithm depends on the initial clustering, best practice demands that prior knowledge be used to inform the choice of the initial clustering and also that the algorithm be ran on a variety of initial clusterings. In the end, only the clustering with the smallest WCSS is retained.


\section*{Appendix D: Fuzzy $K$-means}\label{app_D}
In real data, it often happens that some points do not belong clearly to one cluster or another and it might be desirable to capture this fact. This is what fuzzy clustering algorithms do by associating to a data point, not a cluster label, but a list of weights representing the degrees of belonging to each cluster. In the fuzzy version of $K$-means invented by Dunn \cite{Dunn73}, this is done by attempting to minimize the fuzzy WCSS
$$
fWCSS=\sum_{k=1}^K\sum_{\bm{x}\in \mathcal{C}_k}w(\bm{x},k)^2\|\bm{x}-\bm{c}_k\|^2,
$$
where the weights $w(\bm{x},k)$ are defined as the normalized inverse squared distances to the cluster centroids $\bm{c}_k$, and the centroid $\bm{c}_k$ is defined as a weighted mean of $\mathcal{C}_k$:
$$
w(\bm{x},k) = \frac{\|\bm{x} - \bm{c}_k\|^{-2}}{\sum_{\ell=1}^K\|\bm{x}-\bm{c}_{\ell}\|^{-2}},\quad \bm{c}_k = \frac{\sum_{\bm{x}\in\mathcal{C}_k}w(\bm{x},k)^2\bm{x}}{\sum_{\ell=1}^Kw(\bm{x},\ell)^2}.
$$
For any given $\bm{x}$, the weights $w(\bm{x},k)$ are non negative and they sum to 1. 

As an example, if $K=3$ and $w(\bm{x},1) = 0.47$, $w(\bm{x},2)=0.52$, $w(\bm{x},3)=0.01$, we might say that $\bm{x}$ belongs to cluster $2=\mathrm{argmax}_kw(\bm{x},k)$, but with low degree of confidence as $\bm{x}$ appears to lie somewhat halfway between $\bm{c}_1$ and $\bm{c}_2$.

\end{document}